\documentclass{article}
\usepackage{style}

\usepackage{soul}
\usepackage{times}
\usepackage{hhline}
\usepackage{algorithm}
\usepackage[noend]{algorithmic}
\usepackage{soul}
\usepackage{url}
\usepackage[hidelinks]{hyperref}
\usepackage[utf8]{inputenc}
\usepackage[small]{caption}
\usepackage{graphicx}
\usepackage{amsmath}
\usepackage{multirow}
\usepackage{amsthm}
\usepackage{booktabs}
\usepackage{xspace}
\usepackage{amssymb}
\usepackage{url}
\usepackage{pgf, tikz}
\usetikzlibrary{arrows, automata}
\usepackage{tabularx}
\usepackage{subfig}
\usepackage{xspace}
\usepackage{adjustbox}
\usepackage{subcaption}

\newtheorem{definition}{Definition}
\newtheorem{example}{Example}
\newtheorem{proposition}{Proposition}

\newtheorem{theorem}{Theorem}

\def\styleex{\rm}

\newcommand{\wrt}{w.r.t.\xspace}

\def\+{\mbox{+}}
\def\-{\mbox{-}}
\def\ug{\ensuremath{\mbox{=}}}
\def\agg{q}
\def\<{\langle}
\def\>{\rangle}
\def\styleex{\rm}

\newcommand{\sem}[1]{\ensuremath{\mathtt{#1}}\xspace}

\newcommand{\Set}[1]{\ensuremath{#1}\xspace}

\def\arg{\Set{A}}
\def\att{\Set{R}}
\def\supp{\Set{S}}

\def\goal{\ensuremath{\tt g}}
\newcommand{\ra}[1]{{\ensuremath{\tt a_{#1}}}}
\newcommand{\rs}[1]{{\ensuremath{\tt s_{#1}}}}

\newcommand{\funz}[1]{\textnormal{\ensuremath{\textsf{#1}}}\xspace}
\def\drelu{\funz{DReLU}\xspace}
\def\ddrelu{\funz{dDReLU}\xspace}

\def\QBAF{\ensuremath{\Delta}\xspace}

\def\nAF{AF\xspace}
\def\nQBAF{QBAF\xspace}

\newcommand{\tQBAF}{\ensuremath{\< \arg,\att,\supp,\tau\>}\xspace}

\usepackage{xcolor}
\usepackage{adjustbox}

\title{Double Rectified Linear Unit-based Modular Semantics for\\ Quantitative Bipolar Argumentation Framework}

\author{
Gianvincenzo Alfano\and
Sergio Greco\and
Lucio La Cava\and
Francesco Parisi\and
Irina Trubitsyna\\
\affiliations
Department of Informatics, Modeling, Electronics and System Engineering\\
University of Calabria, Italy\\
\emails
\{g.alfano, greco, lucio.lacava, fparisi, i.trubitsyna\}@dimes.unical.it
}

\begin{document}

\maketitle

\begin{abstract} 
Quantitative Bipolar Argumentation Frameworks (QBAFs) provide an alternative approach to computing argument acceptability in Bipolar Argumentation Frameworks (BAFs). 
Each argument is assigned an initial strength, which is then updated to a final strength by considering the influence of both its attackers and supporters.
Over the years, several semantics have been proposed to compute argument acceptability in {\nQBAF}s, yet they often yield divergent or counterintuitive results, even for simple acyclic cases.
We introduce novel gradual semantics for {\nQBAF}s that address these limitations, producing results that align more closely with intuitive expectations, while satisfying established rationality postulates from the literature.
Furthermore, we study its convergence behavior, proving that it converges not only for acyclic {\nQBAF}s but also for broader classes of cyclic frameworks. 
\end{abstract}

\section{Introduction}\label{sec:intro}

Formal argumentation has become a prominent area within Artificial Intelligence (AI) research~\cite{BenchCapon2007619,Rahwan-Simari-Book,AtkinsonBGHPRST17}. 
At the heart of this field lies Dung's \emph{abstract Argumentation Framework (AF)}, a foundational yet expressive formalism for representing disputes among agents~\cite{Dung95}.
An AF consists of a set of abstract arguments and a binary attack relation that defines how arguments interact. Arguments themselves are treated as abstract entities, whose acceptability depends solely on their position in the attack structure.
This framework can be naturally visualized as a directed graph, where nodes represent arguments and edges denote attacks. 
According to the original proposal, the semantics of \nAF is based on the concept of extensions, i.e., sets of arguments that can be collectively accepted (corresponding to models in classical logic), and of argument acceptance, i.e., a \textit{goal} argument is credulously (resp., skeptically) accepted if it belongs to at least (resp., all) extensions.

To enhance the expressive power, Dung's framework has been extended in several directions.  For the aim of this paper, we recall the \textit{Bipolar Argumentation Framework} (BAF), which adds support relations alongside attacks~\cite{NouiouaR11,Nouioua13,VillataBGT12}, 
and frameworks whose semantics associate a degree of acceptability with each argument, rather than simply saying whether an argument is credulously or skeptically accepted.
In a BAF, there are two kinds of relations: \textit{attacks} and \textit{supports}. Thus, a BAF can be represented by a graph with two types of edges, representing attacks and supports. 
Different interpretations of support have been proposed. 
Regarding the ``necessary'' and ``deductive'' interpretations, which are those that have received more attention, the semantics can be defined in terms of a mapping to an ``equivalent'' AF. 
This means that BAF does not increase the expressivity and complexity of AF, and supports are introduced to simplify the knowledge representation process.
Considering AF-based frameworks where argument acceptance is given by a degree (usually in $[0,1]$), two types of frameworks, which differ in how the acceptance degree is calculated, have attracted the attention of researchers:
\textit{Probabilistic Argumentation Framework} (PAF)~\cite{LiON11,Hunter12} and \textit{Quantitative Bipolar Argumentation Framework} ({\nQBAF})~\cite{cayrol2005graduality,baroni2019fine,amgoud2018gradual,Mossakowski-core}.
In PAF, arguments (or attacks or both) have an associated initial probability value, and the problem consists in assigning to each argument a final value denoting the probability that it is accepted.
In \nQBAF, arguments (or attacks or both) have an associated initial value called \textit{strength}, and the problem consists in assigning to each argument a final strength value by considering the strength values of attackers and supporters.
The main difference between these frameworks is that in PAF the acceptance probability of arguments is computed by:
$(i)$ first assigning a probability to the possible worlds (i.e., AFs obtained by leaving or removing probabilistic arguments), 
$(ii)$ then considering, for each possible world, the extensions (according to the underlying semantics) and assigning them a probability, and, finally, $(iii)$ determining the acceptability value of an argument, considering the probabilities of the extensions containing it.
On the other hand, the semantics of \nQBAF do not rely on the concept of extensions, but the final strength of an argument is computed taking into account the strengths of its attackers and supporters.
An important difference between the two approaches is that computing the argument strength in \nQBAF is very efficient (usually polynomial-time), as we will see in the rest of the paper, while computing the probabilistic acceptance of arguments in PAF can be very hard (up to ${FP}^{\tt\#P}\mbox{-}$complete)~\cite{FazzingaFP15,FazzingaFF19}, since in the general case an exponential number of possible AFs should be considered.

An example of \nQBAF is discussed next.

\begin{figure}[t]
    \centering
    \includegraphics[width=0.7\columnwidth]{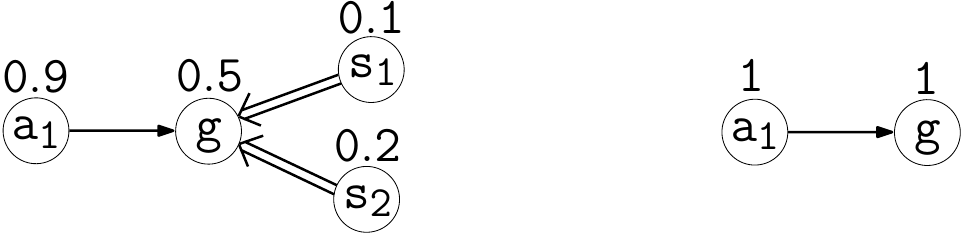}
    \caption{{\nQBAF}s  of Example~\ref{ex:intro1}  (left) and Example~\ref{ex:intro-mlp} (right). Numbers on the top of nodes (i.e., arguments) refer to the initial scores.\vspace*{-3mm}}
    \label{fig:intro1}
\end{figure}

\begin{example}\label{ex:intro1}\styleex
Consider a debate where participants discuss the impact of AI on employment, giving rise to arguments:

\begin{itemize}
    \item[\goal:] ``AI's impact on employment remains uncertain: while automation may replace some tasks, it might enable new roles we can't yet predict'';
    \item[\ra{1}:] ``Automation driven by AI will inevitably cause massive job losses, as it will particularly impact manufacturing and administrative sectors'';
    \item[\rs{1}:] ``Some analysts note that previous technological shifts sometimes led to new forms of work, though it's unclear if this will happen with AI''; 
    \item[\rs{2}:] ``Certain long-term empirical studies suggest that technological progress can complement human labor, but these effects might be modest and take years to appear''.
\end{itemize}

Accordingly, $\ra{1}$ \emph{attacks} $\goal$ by challenging its optimism, while $\rs{1}$ and $\rs{2}$ \emph{support} $\goal$ with historical references and empirical data, respectively. 
We assume initial strengths $\tau_{\goal}=0.5$, $\tau_{\ra{1}}=0.9$, $\tau_{\rs{1}}=0.1$, and $\tau_{\rs{2}}=0.2$.
The resulting \nQBAF is shown in Figure~\ref{fig:intro1}(left),  where initial strengths are reported on top of the corresponding arguments.~\hfill~$\Box$
\end{example}

The semantics of a \nQBAF consists in the way the final strength of arguments is computed. Over time, several \emph{argumentation semantics} for \nQBAF have been introduced~\cite{BesnardH01,CayrolL05,AmgoudBDV17,AmgoudBDDKM18,Potyka19,AmgoudD19,DoderAV23}, including \emph{DF-QuAD} (\sem{dfq})~\cite{RagoKR16}, \emph{Restricted Euler-based} (\sem{eul})~\cite{AmgoudB18Euler}, \emph{Quadratic Energy} (\sem{qen})~\cite{Potyka18QE}, and \emph{MLP} (\sem{mlp})~\cite{Potyka21MLP}. 

A first problem with some existing \nQBAF semantics is that they can give very different results (in some cases, even opposite).
Indeed, considering the \nQBAF of Example~\ref{ex:intro1}, the final strength of the goal argument \goal \ is $\tt 0.62$ (resp., $\tt 0.41$, $\tt 0.19$, and $\tt 0.37$) under \sem{mlp} (resp., \sem{eul}, \sem{dfq}, \sem{qen}). 
A second, though not less important, issue concerning \nQBAF semantics is that they often provide unintuitive results, as shown in Section~\ref{sec:prel} and anticipated next for one of the semantics analyzed next.

\begin{example}\label{ex:intro-mlp}\styleex
Consider a \nQBAF obtained from that of Figure~\ref{fig:intro1}(left) by: $i)$ removing arguments $\rs{1}$ and $\rs{2}$ and relations involving them (i.e., supports from $\rs{1}$ and $\rs{2}$ to $\goal$), and $ii)$ set $\tt \tau_{\ra{1}}=\tau_{\goal}= 1$.
The resulting \nQBAF is shown in Figure~\ref{fig:intro1}(right).
Under \sem{mlp} semantics, the final strength of $\goal$ is $\tt 1$ independently of the initial strength $\tau_{\ra{1}}$ of its attacker ${\ra{1}}$.~\hspace*{\fill}$\Box$
\end{example}

Regarding the previous example, as it will be clarified in what follows, the novel semantics proposed in this paper assign a final strength of $\tt 0.5$ to argument  $\goal$.
This value intuitively captures the fact that the acceptability of argument $\goal$ is computed by equally balancing its initial strength $\tt \tau_{\goal}$ with that of its attacker (i.e., $\tau_{\ra{1}}$). 

Although certain existing semantics also address this issue by yielding a final value of 0.5, they nonetheless fail to resolve other problems, as it will be discussed in Section~\ref{sec:novel-sem}.

\vspace*{2mm}
\noindent
\textbf{Contributions.}
Our main contributions are as follows. 
\begin{itemize}
\item 
After showing that existing \nQBAF semantics can yield counterintuitive results, even in the acyclic case, we propose novel gradual semantics that solve those issues.
\item 
We compare the proposed semantics with the existing ones with respect to well-established postulates, showing that they satisfy foundational principles established in the literature.

\item 
To further guide the user to distinguish between the proposed semantics, we compare them empirically. 
That is, we have conducted an experimental analysis aimed at evaluating the output (i.e., the final score) on different classes of graphs. 

\item 
We investigate the convergence behavior of the proposed semantics, establishing that they converge not only in the acyclic case but also within broader classes of {\nQBAF}s.
This has also been empirically supported by implementing them in the Attractor Library~\cite{Potyka22}.
\end{itemize}

\section{Preliminaries}\label{sec:prel}

In this section, we first review the syntax of \nQBAF, and then recall well-known gradual semantics. 

\begin{definition}[Syntax]\label{def:QBAF}
A Quantitative Bipolar Argumentation Framework (\nQBAF) is a quadruple $\QBAF=\tQBAF$ consisting of a set of arguments $\arg$, an attack relation $\att\subseteq \arg\times\arg$, a support relation $\supp\subseteq \arg\times\arg$, and a total function $\tau: \arg\rightarrow [0,1]$ that assigns the initial strength $\tau(a)$ (also denoted as $\tau_a$) to every argument $a\in\arg$.    
\end{definition}

Thus, a \nQBAF can be represented by a signed, directed graph (or simply graph) $\<\arg, \att, \supp\>$ with two types of edges of the form $a \rightarrow b$, representing the relation $a$ attacks $b$ (i.e., $(a,b) \in \att$), and $a \Rightarrow b$, representing the relation $a$ supports $b$ {(i.e., $(a,b) \in \supp$)}.
Intuitively, attacking arguments contribute to decreasing the strength of an attacked argument, while supporting arguments contribute to increasing the strength of a supported argument.
Since the strength of arguments is a real value in the range $[0,1]$, a strength $0$ corresponds to the total rejection of an argument, $1$ to the total acceptance, and values between $0$ and $1$ express various degrees of acceptance.

We say that argument $b\in \arg$ is \textit{reachable} from $a\in\arg$ w.r.t. \nQBAF $\QBAF=\tQBAF$, iff there is a path from $a$ to $b$ in the graph $\<\arg,\att,\supp\>$. 
A \nQBAF $\QBAF$ is said to be \textit{acyclic} iff the graph $\<\arg,\att,\supp\>$ does not contain cycles.

Several gradual semantics have been proposed so far to compute the acceptance of arguments. 
As said before, a gradual semantics for a \nQBAF $\QBAF$ is the way a total function $\rho_\QBAF:\arg\rightarrow[0,1]$, said \emph{update function}, assigns a value in $[0,1]$, called \textit{(final) strength}, to arguments. 
\nQBAF semantics are typically classified within the family of \textit{modular semantics}~\cite{Mossakowski-core}. 
That is, the final strength $\rho_\QBAF(a)$ (or simply $\rho(a)$ whenever the \nQBAF \QBAF is clear from the context) of any argument $a\in\arg$ is iteratively computed, starting from the initial one $\tau_a$, by considering the strength $\rho(b)$ of each attacker/supporter $b$ of $a$.
The name modular is due to the fact that the function $\rho$ consists of the composition of two functions: 
$i)$ an \textit{aggregation function} $\alpha(a)$ which combines the strengths of $a$'s attackers and supporters; and  
$ii)$ an \textit{influence function} $\iota(\tau_a,\alpha(a))$, which adjusts the strengths according to the initial strength $\tau_a$ and the aggregation $\alpha(a)$. We use $\rho(a)=\iota(\tau_a,\alpha(a))$ to denote the composition of the influence and aggregation function.

The iterative procedure for computing the final strengths of any acyclic \nQBAF $\QBAF=\tQBAF$ under modular semantics is equivalent to a forward pass following a topological linear ordering of the arguments, allowing the final strength to be computed in linear time~\cite{Potyka19}. 

The domain of argumentation semantics encompasses a large and continually expanding landscape of proposed solutions. 
Given the large number of existing approaches, we have focused on four specific semantics that have, so far, received significant attention from the community and serve as representative examples of distinct computational properties. 
We reserve the examination of other relevant modular semantics for future work. 
For each semantics, we present the formula computing the final strength and an example that highlights some of its weaknesses.

\paragraph{Multi Layer Perceptrons-based.} 
The \textit{MLP-based} semantics (\sem{mlp})~\cite{Potyka21MLP} uses the so-called \textit{sum-based} aggregation function, which is formally stated in the next equation.

\begin{equation}\label{eq:sum}
\alpha(a)=\sum\limits_{(b,a)\in\supp}\rho(b)-\sum\limits_{(b,a)\in\att}\rho(b)
\end{equation}

\noindent 
while the update function gives:
\begin{equation}\label{eq:mlp}
\rho(a)= \sigma\left(ln\left(\frac{\tau_a}{1-\tau_a}\right)+\alpha(a)\right)
\end{equation}

\noindent
where $\sigma(z)=\frac{1}{1+e^{-z}}$ is the standard sigmoid function, and we assume the standard convention that $ln(0)=-\infty$.

\begin{example}\label{ex:prel-mlp}\styleex
Consider the \nQBAF $\QBAF_{\ref{ex:intro-mlp}} =\tt \< \{\ra{1},\goal\},$ $\tt \{\tt (\ra{1},\goal)\},$ $\tt \emptyset, \{\tau_{\ra{1}}=\tau_{\goal}=1\} \>$ discussed in Example~\ref{ex:intro-mlp} (see Figure~\ref{fig:intro1}, right).
By applying Equation~\ref{eq:sum}, we have that $\alpha(\goal)= -1$, and, from Equation~\ref{eq:mlp}, that $\rho(\goal) = \sigma(\infty)=\tt 1$.~\hspace*{\fill}$\Box$
\end{example}

The result of the previous example is unexpected because, being $\ra{1}$ not attacked, $\rho({\ra{1}}) = \tau_{\ra{1}} = 1$ and, therefore, the attack from $\ra{1}$ to $\goal$ should have decreased the strength of $\goal$.

\paragraph{Restricted Euler Based.}
The \textit{Restricted Euler Based} semantics (\sem{reb})~\cite{AmgoudB18Euler} uses the sum aggregation function (cf. Equation~\ref{eq:sum}) and the following update  function:
\begin{equation}\label{eq:reb}
\rho(a)=
1 - \frac{1 - \tau_a^2}{1 + \tau_a\cdot e^{\alpha(a)}}
\end{equation} 
 
\begin{example}\label{ex:prel-REB}\styleex
Consider again the \nQBAF $\QBAF_{\ref{ex:intro-mlp}}$. 
By applying Equation~\ref{eq:sum} we have that $\alpha(\goal)= {-}1$, and, from Equation~\ref{eq:reb}, that $\rho(\goal) =\tt 1$.~\hspace*{\fill}$\Box$
\end{example}

As observed for MLP-based semantics, this semantics also does not capture the expected behavior.

\begin{figure}[t]
    \centering
    \includegraphics[width=0.7\columnwidth]{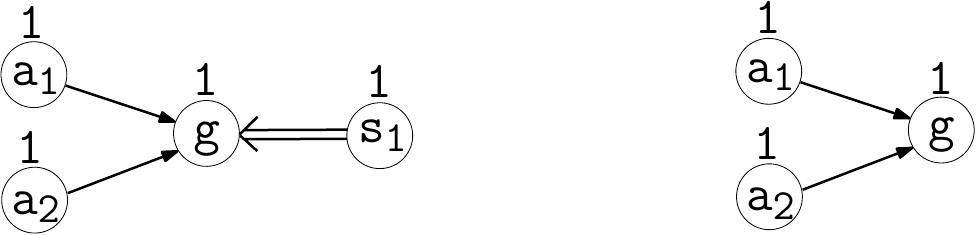}
    \caption{(From left) {\nQBAF}s $\QBAF_{\ref{ex:prel-DFQUAD}}$ and $\QBAF_{\ref{ex:prel-QE}}$ of Examples~\ref{ex:prel-DFQUAD} and~\ref{ex:prel-QE}.}
    \label{fig:prel}
\end{figure}

\paragraph{DF-QuAD.}
The aggregation and update functions of \textit{Discontinuity-free Quantitative Argumentation Debate} semantics (\sem{dfq})~\cite{RagoKR16}, where $\pi$ is used (instead of $\alpha$) to distinguish from the sum-based aggregation, are:

\vspace*{-2mm}

\begin{equation}\label{eq:prod}
{\pi}(a) = \prod\limits_{(b,a) \in \att}(1-\rho(b)) - 
            \prod\limits_{(b,a) \in \supp}(1-\rho(b))
\end{equation}
\begin{equation}\label{eq:dfq}
\rho(a)=
\begin{cases}
\tau_a\cdot (1 + {\pi}(a))      & \text{if } {\pi}(a) \le 0 \\
\tau_a\cdot (1 - {\pi}(a)) + {\pi}(a) & \text{if } {\pi}(a) > 0
\end{cases}
\end{equation}

\begin{example}\label{ex:prel-DFQUAD}\styleex
Consider again the \nQBAF $\QBAF_{\ref{ex:intro-mlp}}$. 
By applying Equation~\ref{eq:dfq}, as ${\pi}({\goal}) \ug -1$ (as the product over an empty set, particularly $\supp$, is $1$), we have that  $\rho(\goal) = \tt 1\cdot (1+(-1))= 0$.

Consider now the \nQBAF $\QBAF_{\ref{ex:prel-DFQUAD}} =\<\arg={\tt\{\ra{1},\ra{2},\goal,}{\tt\rs{1}\},} 
\{\tau_x =1\ \forall x\in\arg\}, \tt \{(\ra{1},\goal), (\ra{2},\goal)\}, \{(\rs{1},\goal)\} \>$, obtained from $\QBAF_{\ref{ex:intro-mlp}}$ by adding the argument $\ra{2}$ attacking $\goal$ and the argument $\rs{1}$ supporting $\goal$; both the new arguments have initial strength equal to $\tt 1$ (see Figure~\ref{fig:prel}, left).
By applying Equation~\ref{eq:dfq}, as ${\pi}({\goal})=0$, we have that $\rho(\goal) = \tt\tau_{\goal}= 1$.~\hspace*{\fill}$\Box$
\end{example}

Both the results for the {\nQBAF}s $\QBAF_{\ref{ex:intro-mlp}}$ and $\QBAF_{\ref{ex:prel-DFQUAD}}$ are counterintuitive. In the first case, the final strength of the argument $\tt a$ is $\tt 0$, regardless of its initial strength.
In the second case, after adding an attack and a support of the same magnitude to the previous framework, we obtain a completely different result.

\paragraph{Quadratic Energy.}
The \textit{Quadratic Energy} semantics~\cite{Potyka18QE} (\sem{qen}) uses the same (sum) aggregation function as per \sem{mlp} and \sem{reb} (cf. Equation~\ref{eq:sum}), while the {update} function is as follows:
\begin{equation}\label{eq:qen}
\rho(a)=
\begin{cases}
\left(1 - E(a)\right)\cdot \tau_a  & \text{if } {\alpha(a)} \le 0 \\
E(a) + \left(1 - E(a)\right)\cdot \tau_a  & \text{if } {\alpha(a)} > 0 
\end{cases}
\end{equation}
\noindent 
where 
$E(a) = \frac{\alpha(a)^2}{1 + \alpha(a)^2}$.

\begin{example}\label{ex:prel-QE}\styleex
Consider again the \nQBAF $\QBAF_{\ref{ex:intro-mlp}}$. 
As $\alpha(\goal) \mbox{=} \tt -1$, we have that $E(\goal) = 0.5$ and then $\rho(\goal) = ({\tt 1}- $ $E({\goal}))\cdot \tau_{\goal}=\tt 0.5$.

Consider now the \nQBAF $\QBAF_{\ref{ex:prel-QE}} =\tt  \<\{\ra{1},\ra{2},\goal\},$ $\tt \{(\ra{1},\goal),$ $\tt (\ra{2},\goal)\},$ $\tt \emptyset, \{\tau_{\ra{1}} =\tau_{\ra{2}} =\tau_{\goal} = 1\} \>$, that is obtained from the \nQBAF $\QBAF_{\ref{ex:intro-mlp}}$ by adding another argument ($\ra{2}$) with initial strength $\tau_{\ra{2}}=1$ and the attack $(\tt \ra{2},\goal)$ (see Figure~\ref{fig:prel}, right).
As $\alpha(\goal) = \tt -2$, we have that $E(\goal) = \tt 0.8$ and $\rho(\goal) = 0.2$.~\hfill $\Box$
\end{example}

Semantics \sem{qen} appears to capture the intuitive semantics of both the $\QBAF_{\ref{ex:intro-mlp}}$ and $\QBAF_{\ref{ex:prel-QE}}$ frameworks.
However, as we will see in the next section, in some cases the results are, in our opinion, unsatisfactory.
It is important to note that all the examples used to highlight weaknesses regard acyclic frameworks.

\section{Novel Semantics for \nQBAF}\label{sec:novel-sem}
The semantics reviewed in the previous section have been compared at the level of principles~\cite{AmgoudB18Euler,Potyka18QE,Mossakowski-core} and, as will be shown in Table \ref{tab:semantics-principles}, all of them satisfy most of the principles defined in the literature. 
However, as previously illustrated, most of them may produce unintuitive outcomes, even for trivial, acyclic {\nQBAF}s. 
For instance, \sem{mlp} and \sem{reb} semantics are not intuitive when the initial strength of an argument is equal to $1$, as its final value is not influenced by other arguments attacking or supporting it. 
Intuitively, for the framework $\QBAF_{\ref{ex:intro-mlp}}$ of Examples~\ref{ex:prel-mlp} and \ref{ex:prel-REB}, we would have expected the acceptability of argument $\goal$ to decrease when $\goal$ receives only attacks from unattacked arguments.

For \sem{dfq} semantics, we point out two problems. First, in computing the final score of an argument $x$, it may fail to consider the initial $x$'s strength. 
Considering, e.g., the \nQBAF $\QBAF_{\ref{ex:intro-mlp}}$ we have that $\rho(\goal) = 0$ regardless of the value $\tau_{\goal}$.
Second, in the presence of attackers and supporters with scores $1$, regardless of their numbers, the aggregation function $\pi$ returns $0$ and, therefore, the final and initial strength of the attacked argument coincide, as it holds for argument $\goal$ in \nQBAF $\QBAF_{\ref{ex:prel-DFQUAD}}$. 

Concerning \sem{qen} semantics, it overcomes the previous issues but still fails on a closely related problem, as shown next.

\begin{figure}[t!]
    \centering
    
    \begin{minipage}{0.45\columnwidth}
        \centering    \includegraphics[width=0.7\columnwidth]{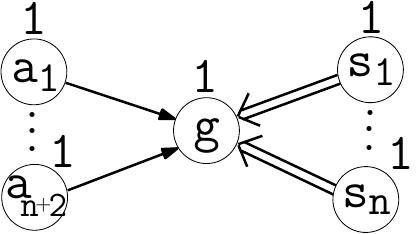}
    \end{minipage}
    \hfill
    \begin{minipage}{0.45\columnwidth}
        \centering
   \includegraphics[width=0.6\columnwidth]{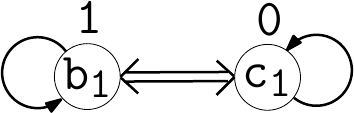}
    \end{minipage}
    \caption{{\nQBAF} $\QBAF_{\ref{ex:unintuitive-QE}}$ of Examples~\ref{ex:unintuitive-QE} and~\ref{ex:unintuitive-QE-cont} (left), and \nQBAF $\QBAF$ discussed in the proof of Proposition~\ref{prop:divergence-general} (right).}
    \label{fig:unintuitive}
\end{figure}

\begin{example}\label{ex:unintuitive-QE}\styleex
Consider the \nQBAF $\QBAF_{\ref{ex:unintuitive-QE}}$ shown in Figure~\ref{fig:unintuitive} (left), obtained from $\QBAF_{\ref{ex:prel-QE}}$ by adding $n$ arguments $\rs{1},...\rs{n}$ supporting $\goal$, $n$ further arguments $\ra{3},...,\ra{n+2}$ attacking $\goal$, and $\tau_x=1$ for each argument $x$ in $\arg$. 
As for $\QBAF_{\ref{ex:prel-QE}}$,  we still have that $\alpha({\goal}) = \tt -2$ and, therefore, $\rho({\goal}) = 0.2$.

However, when $n$ is large,  the arguments are nearly evenly split between those supporting and those attacking 
\goal---nevertheless, $\rho({\goal})$ fails to capture this equilibrium.~\hfill $\Box$
\end{example}

The problem evidenced in the \sem{qen} semantics is, of course, present in all semantics using the sum-based aggregation function (cf. Equation~\ref{eq:sum}).

Motivated by these counterexamples, we propose novel semantics, not affected by above discussed issues.

\subsection{Rethinking the Aggregation Function}\label{sec:reth}

We next introduce a variation of the aggregation function that also measures, for any argument $a$, the impact of sum-based aggregation $\alpha(a)$ w.r.t. those of $a$'s supporters and attackers,  respectively denoted as $\alpha^+(a)=\sum_{(b,a)\in\supp}\rho(b)$ and $\alpha^-(a)=\sum_{(b,a)\in\att}\rho(b)$. Formally: 
\vspace*{-1.5mm}
\begin{equation}\label{eq:beta}
\delta_{\agg}(a) \!=\! 
\begin{cases}
\hspace*{8mm}0 & \mbox{if } \alpha^+\!(a) \!=\!\alpha^-\!(a) \!=\! 0 \\
    \frac{\alpha(a) \cdot |\alpha(a)|}{\agg(\alpha^+\!(a),\alpha^-\!(a))} &  \mbox{otherwise.}
\end{cases}
\end{equation}
\noindent
where, let $\agg \in \{\mathrm{max},\mathrm{sum}\}$ be a function that takes the maximum or the sum of two values.
Note that, $\delta_q(a) \leq \alpha(a)$ as $|\alpha(a)| / q(\alpha^+(a),\alpha^-(a)) \leq 1$ and that $\delta_{\mathrm{sum}}(a) \leq \delta_{\mathrm{max}}(a)$, as $(\alpha^+(a)+\alpha^-(a)) \geq \mathrm{max}(\alpha^+(a),\alpha^-(a))$.

\begin{example}\label{ex:beta}\styleex
Considering the \nQBAF $\QBAF_{\ref{ex:unintuitive-QE}}$ of Example~\ref{ex:unintuitive-QE}, we have that $\delta_{\mathrm{sum}}(\goal) = -2/(n+1)$ and 
$\delta_{\mathrm{max}}(\goal) = -4/(n+2)$.
As a consequence, we have that for $n=0$ we get the values $\delta_{\mathrm{sum}}(\goal) = \delta_{\mathrm{max}}(\goal) = -2$, while
for $n=1$ (resp., $n=5$) we get the values $\delta_{\mathrm{sum}}(\goal) = -1$ and $\delta_{\mathrm{max}}(\goal) = -4/3$ (resp., $\delta_{\mathrm{sum}}(\goal) = -1/3$ and $\delta_{\mathrm{max}}(\goal) = -4/7$).
\hfill$\Box$    
\end{example}

Clearly, $sign(\delta_q(a)) {=} sign(\alpha(a))$ for any $q \in \{\mathrm{max},$ $\mathrm{sum}\}$.
The next proposition states that $\delta_q$ increases \wrt $\alpha$.

\begin{proposition}\label{prop:alpha}
Given a \nQBAF $\tQBAF$ and two arguments $a,b\in\arg$ such that 
$\alpha(a) \geq \alpha(b)$, then $\delta_q(a) \geq \delta_q(b)$ holds for any $q\in\{\mathrm{sum}, \mathrm{max}\}$.
\end{proposition}

The first semantics we introduce, called \textit{modified Quadratic Energy} (\sem{mqe}), is derived from \sem{qen} by replacing the aggregation function $\alpha$ with $\delta_q$.
As $\agg \in \{\mathrm{max},\mathrm{sum}\}$, there are two different ways to get the final strength $\rho_q$. 

\begin{definition}[\sem{mqe}]
For any \nQBAF $\QBAF=\tQBAF$ the strength $\rho_{\agg}(a)$ of an argument $a \in \arg$, where $\agg \in \{\mathrm{max},\mathrm{sum}\}$, is obtained as follows:
\begin{equation}\label{eq:mqe}
\rho_{\agg}(a)=
\begin{cases}
\left(1 - E_{\agg}(a)\right)\cdot \tau_a  & \text{if } {\delta_{\agg}(a)} \le 0 \\
E_{\agg}(a) + \left(1 - E_{\agg}(a)\right)\cdot \tau_a  & \text{if } {\delta_{\agg}(a)} > 0 
\end{cases}
\end{equation}
$$ 
\mbox{where} \hspace*{9mm}
E_{\agg}(a) = \frac{(\delta_{\agg}(a))^2}{1 +( \delta_{\agg}(a))^2}.
\hspace*{32mm}
$$
\end{definition}

Semantics \sem{mqe} fixes the issue of Example \ref{ex:unintuitive-QE}, as shown next. It allows for choosing the parameters $\mathrm{max}$ and $\mathrm{min}$ in the calculation of $\delta_{\agg}(a)$, according to the user's preference.

\begin{example}\label{ex:unintuitive-QE-cont}\styleex
Continuing with Example \ref{ex:beta}, under the \sem{mqe} semantics
we have that for $n=0$ we get the values $\rho_{\mathrm{sum}}(\goal) = \rho_{\mathrm{max}}(\goal) = 0.2$, while
for $n=1$ (resp., $n=5$) we get the values $\rho_{\mathrm{sum}}(\goal) = 0.5$ and $\rho_{\mathrm{max}}(\goal) = 0.36$ (resp., $\rho_{\mathrm{sum}}(\goal) = 0.9$ and $\rho_{\mathrm{max}}(\goal) \approx 0.75$. 
Thus, for $n=0$ we have that the argument $\goal$ is almost rejected, while for $n=5$ it is almost accepted.~\hfill~$\Box$
\end{example}

Exploring alternative definitions for $\sem{mqe}$, such as e.g. replacing $\delta_{\agg}(a)^2$ with $|\delta_{\agg}(a)|$ is reserved for future work.

\subsection{An Alternative Influence Function}

We introduce the \textit{Double Rectified Linear Unit} function, that is defined as $\drelu(z) = \mathrm{max}\{-1,\mathrm{min}\{1,z\}\}$. Basically, it is a piecewise-defined function and is not differentiable in the whole domain (it is not differentiable for $z=\pm 1$), and it can also be defined as follows:
\vspace*{-1mm}
\begin{equation}\label{eq:drelu}
\drelu(z) =   
\begin{cases}
     -1 & \mbox{if }\ z \leq -1 \\
      z & \mbox{if }\ -1 \leq z \leq 1 \\
      1 & \mbox{if }\ z \geq 1     
\end{cases}
\end{equation}
 
The function \drelu is an adaptation of the classical \funz{ReLU} function, but it is also similar to the hyperbolic tangent function, as for large positive values of $z$ it saturates to $1$, whereas for large negative values of $z$ it saturates to $-1$.\footnote{It is also known as \textit{saturating linear transfer function} \cite{CHANG200786} and, for ML experts, it is equivalent to the \textit{clamping} function, i.e., ${\drelu}(z) = \funz{clamp}(z,-1,1)$.}

The next definition introduces a new way of computing the final scores of arguments, called \sem{drl} semantics.

\begin{definition}[\sem{drl}]\label{def:oursacyclic}
For any \nQBAF $\QBAF=\tQBAF$  and constant $\gamma \in \mathbb{R}_{\geq 0}$, the strength $\rho_{\agg}(a)$ of an argument $a \in \arg$, where $\agg \in \{\mathrm{max},\mathrm{sum}\}$, is obtained as follows: 
\begin{equation}\label{eq:oursacyclic}
\rho_{\agg}(a) \!=\! 
\frac{\!\drelu\left((2\tau_a\!-\!1) +  \delta_{\agg}(a) \cdot {\gamma} \right)+1}{2}.
\end{equation}
\end{definition}

In the previous definition, the parameter $\gamma$ is used to assign weights to attackers and supporters with respect to the intrinsic strength $\tau_a$, whose weight is by default equal to $1$.
Giving to $\gamma$ a high value means that the strength of argument $a$ is mainly determined by the status of attackers and supporters, whereas, giving to $\gamma$ a low value, means that the strength of $a$ is mainly influenced by the initial strength $\tau_a$.
For $\gamma = {1}$, we have that $\rho_{\agg}(a) = f^{\mbox{-}1}(\drelu(f(\tau_a)+\delta_{\agg}(a)))$, where $f(\tau_a) = 2\tau_a\mbox{-}1$ stretches the value of $\tau_a$ from range [0,1] to $[\mbox{-}1,1]$, whereas the inverse function $f^{\mbox{-}1}(z) = (z+1)/2$ compresses the value of $z \in [\mbox{-}1,1]$ to be in the range $[0,1]$.
The reason for stretching $\tau_a$ to be in $[\mbox{-1},1]$ (and then compressing the  result) is to homogenize its value with that of $\delta_{\agg}(a) \in (\-\infty,\infty)$. 
Thus, if $\delta_{\agg}(a) = 0$, we have that $\rho_{\agg}(a)=\tau_a$. 
Moreover, assuming $\gamma = {1}$,  for $\delta_{\agg}(a) = -\tau_a$, we have $\rho_{\agg}(a) = \tau_a/2$, i.e., the acceptability of $a$ is halved, whereas for 
$\delta_{\agg}(a) \leq -2 \tau_a$ (resp.,  $\delta_{\agg}(a) \geq 2(1-\tau_a)$, we get $\rho_{\agg}(a) = 0$ (resp., $\rho_{\agg}(a)= 1$).
Even under \sem{drl} semantics, there are two alternative ways to calculate the final strength (as $q \in \{ \mathrm{max},\mathrm{sum}\}$), based on user preferences.

\begin{example}\label{ex:drelu-discrete-1}\styleex 
Consider the \nQBAF $\QBAF_{\ref{ex:unintuitive-QE-cont}}$ of Example \ref{ex:unintuitive-QE-cont}. 
The final strengths of $\goal$ under {\sem{drl}} semantics are   
$\rho_{\mathrm{max}}({\goal}) =\tt 0.833$ and   $\rho_{\mathrm{sum}}({\goal}) =\tt 0.909$.~\hfill $\Box$
\end{example}

\section{Theoretical and Empirical Comparisons}

We next compare the proposed semantics \sem{drl} and \sem{mqe} against the existing ones, both formally (i.e., by looking at axiomatic properties) and empirically.

\subsection{Axiomatic Properties}~\label{sec:principles}
The formal properties of gradual evaluation methods in argumentation have gained significant interest, with various studies defining principles to deepen the foundational understanding and comparison of semantics. In this section, we evaluate our proposed semantics against the principles established in~\cite{AmgoudB18Euler,Potyka18QE,Mossakowski-core}.

A semantics satisfies a postulate if the condition prescribed by the postulate is satisfied by every \nQBAF.
We next recall a list of postulates introduced in the literature~\cite{AmgoudB18Euler,Potyka19}. 
For each of them, we discuss the intuitive meaning. 

\textit{Anonymity} states that isomorphic {\nQBAF}s yield the same results (i.e., final strengths). 
\textit{Independence} says that the strength of arguments is independent of disconnected arguments. \textit{Directionality} states that the strength of an argument depends only on its predecessors in the graph. 
\textit{Equivalence}, also called \textit{relative balance} in \cite{KR2025-GradualABA}, demands that if two arguments have the same initial strengths and equally strong attackers and supporters, then they should have the same final strength. 
\textit{Stability} says that the final strength of arguments having neither attackers nor supporters should be equal to  the initial strength.
\textit{Neutrality} demands that arguments with strength $0$ do not have any impact. 
\textit{Monotonicity}, also called \textit{relative monotonicity} in \cite{KR2025-GradualABA}, states that if argument $a$ is $i)$ attacked by a subset of the attackers of argument $b$, $ii)$ supported by a superset of the supporters of $b$, and $iii)$ the initial strength of $a$ is equal to or greater than that of $b$, then $a$'s strength must be equal to or greater than $b$'s strength. 
\textit{Reinforcement} differs from monotony by considering the case where the two arguments differ only in one attacker and one supporter. 
\textit{Weakening} says that the final strength of arguments that are strictly weaker supported than attacked must be smaller than the initial weight. 
\textit{Strengthening} is dual to weakening and demands that for arguments that are strictly weaker attacked than supported, the strength must be greater than the initial weight. 
\textit{Duality} demands that attacks and supports are treated equally.
\textit{Open-mindedness} says that semantics should be able to move the strength values arbitrarily close (not necessarily equal) to the extreme values $0$ or $1$ if sufficient evidence against or for the argument is given.

\begin{table}[t!]
\centering
\renewcommand{\arraystretch}{0.85} 
\setlength{\tabcolsep}{0pt} 
\begin{tabular*}{\columnwidth}{@{\extracolsep{\fill}} l cccccccccccc @{}}
\toprule
\textbf{~} & \textbf{An} & \textbf{In} & \textbf{Di} & \textbf{Eq} & \textbf{St} & \textbf{Ne} & \textbf{Mo} & \textbf{Re} & \textbf{We} & \textbf{St} & \textbf{Du} & \textbf{Op} \\
\midrule
$\sem{dfq}$ & $\checkmark$ & $\checkmark$ & $\checkmark$ & $\checkmark$ & $\checkmark$ & $\checkmark$ & $\times$ & $\times$ & $\times$ & $\times$ & $\checkmark$ & $\times$ \\
$\sem{reb}$ & $\checkmark$ & $\checkmark$ & $\checkmark$ & $\checkmark$ & $\checkmark$ & $\checkmark$ & $\checkmark$ & $\checkmark$ & $\checkmark$ & $\checkmark$ & $\times$ & $\times$ \\
$\sem{qen}$ & $\checkmark$ & $\checkmark$ & $\checkmark$ & $\checkmark$ & $\checkmark$ & $\checkmark$ & $\checkmark$ & $\checkmark$ & $\checkmark$ & $\checkmark$ & $\checkmark$ & $\checkmark$ \\
$\sem{mlp}$ & $\checkmark$ & $\checkmark$ & $\checkmark$ & $\checkmark$ & $\checkmark$ & $\checkmark$ & $\checkmark$ & $\checkmark$ & $\checkmark$ & $\checkmark$ & $\checkmark$ & $\times$ \\
\midrule
$\sem{mqe}$ & $\checkmark$ & $\checkmark$ & $\checkmark$ & $\checkmark$ & $\checkmark$ & $\checkmark$ & $\checkmark$ & $\checkmark$ & $\checkmark$ & $\checkmark$ & $\checkmark$ & $\checkmark$ \\
$\sem{drl}$ & $\checkmark$ & $\checkmark$ & $\checkmark$ & $\checkmark$ & $\checkmark$ & $\checkmark$ & $\checkmark$ & $\checkmark$ & $\checkmark$ & $\checkmark$ & $\checkmark$ & $\checkmark$ \\
\bottomrule
\end{tabular*}
\caption{Satisfaction of principles {(two-letter abbreviations)} by different semantics. Novel results refer to semantics $\sem{mqe}$ and $\sem{drl}$. \vspace*{-4mm}}
\label{tab:semantics-principles}
\end{table}

As stated next and summarized in Table~\ref{tab:semantics-principles}, our proposed semantics satisfy all the above-mentioned principles.

\begin{proposition}\label{prop:principles}
    Semantics \sem{mqe} and \sem{drl} satisfy Anonymity, Independence, Directionality, Equivalence, Stability, Neutrality, Monotony, Reinforcement, Weakening, Strengthening, Duality, and Open-mindedness, for any $\agg\in\{\mathrm{sum},\mathrm{max}\}$.
\end{proposition}

\subsection{Empirical  Comparison}\label{subsec:practical}

From the axiomatic characterization (cf. Proposition~\ref{prop:principles}), it follows that $\sem{mqe}$ and $\sem{drl}$ adhere to the same principles. 
To further compare them from an empirical perspective, we designed experiments intended to evaluate the differences between these semantics, e.g. when $\delta_{\mathrm{sum}}$ is preferable to $\delta_{\mathrm{max}}$, or when $\sem{drl}$ is preferable to $\sem{mqe}$. 
\\
For each $n$$\in$$\{n_0\mbox{=} 1, n_1 \mbox{=} 2, n_2 \mbox{=} 5, n_3 \mbox{=} 10, n_4 \mbox{=} 100,n_5\mbox{=}1000\}$, we have generated $100$  
{\nQBAF}s  ${\cal D}_{n}=\{\QBAF_n^1,\dots,\QBAF_n^{100}\}$ of the form shown in Figure~\ref{fig:unintuitive} (left) where $n$ denotes the number of supporters (i.e., parameter $n$ in the figure).
The initial strength of arguments, differently from the \nQBAF shown in Figure~\ref{fig:unintuitive} (left) where it has been fixed to $1$ for all arguments, has been randomly assigned (thus it might differ for distinct arguments), still guaranteeing that $\alpha(\goal)$ approximates $\-2$.  

The construction starts with ${\cal D}_{n_0}$ and each ${\cal D}_{n_j}$, with ${j>0}$, is obtained by augmenting {\nQBAF}s from ${{\cal D}_{n_{j\text{-}1}}}$ with new attackers and supporters, without changing the values of initial strengths for the arguments present in the QBAF{s} of ${\cal D}_{n_{j\text{-}1}}$.
The resulting dataset ${\cal D}\mbox{=}{\{{\cal D}_{n_0},..,{\cal D}_{n_5}\}}$ contains ${600}$ {\nQBAF}s. 

\begin{figure}[t!]
    \centering
    \begin{minipage}{0.62\columnwidth}
        \centering
        \includegraphics[width=\linewidth]{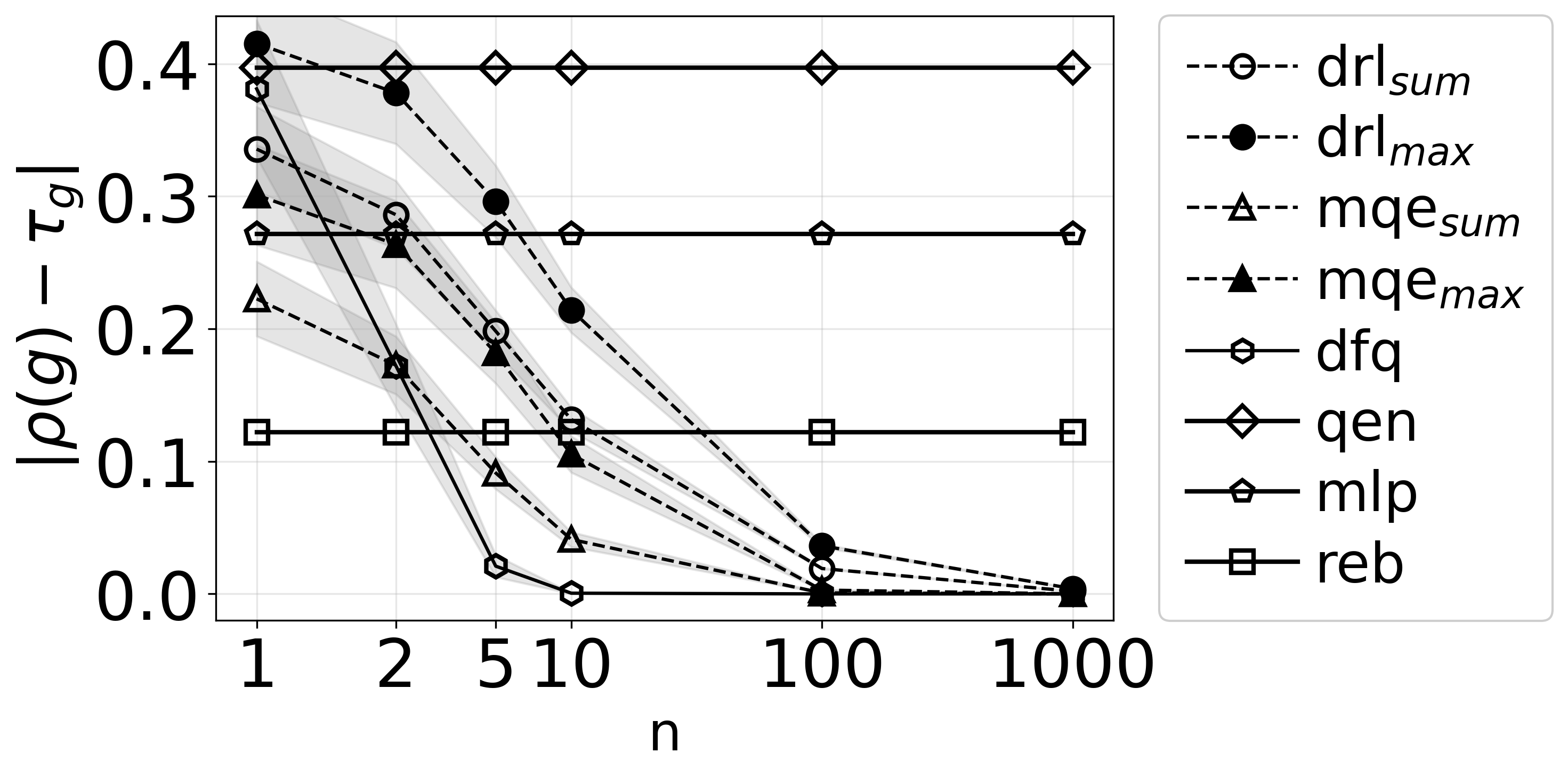}
    \end{minipage}\hfill
    \begin{minipage}{0.38\columnwidth}
        \centering
        \includegraphics[width=0.9\linewidth]{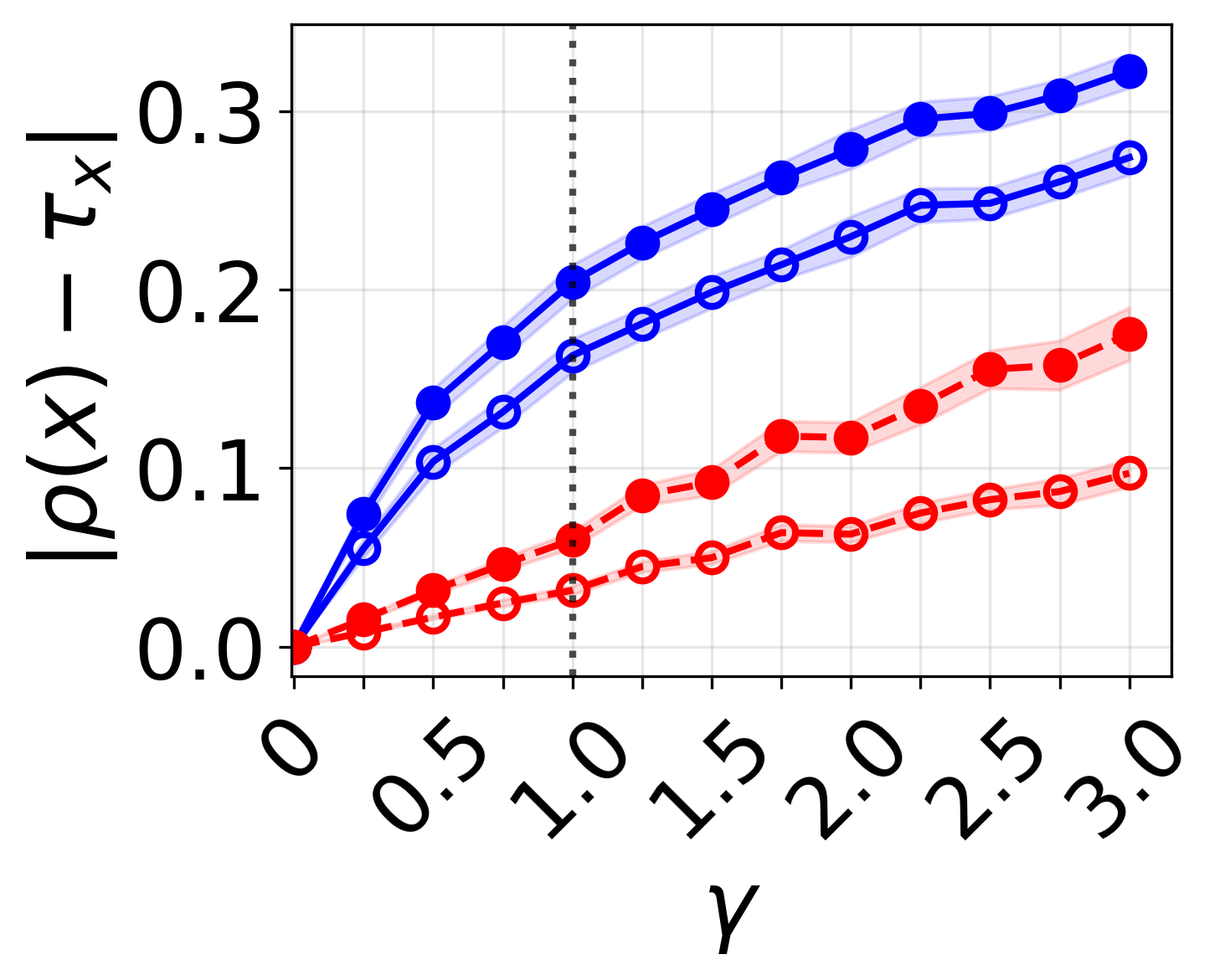}
    \end{minipage}
    \vspace*{-3mm}\caption{(Left) Average values of $|\rho(\goal)-\tau_{\goal}|$ under different semantics for the {\nQBAF}s of ${\cal D}_n$, with  $n \in \{1,2,5,10,100,1000\}$. 
    (Right) Average values in $|\rho(x)-\tau_{x}|$ (averaged over all arguments $x$) under semantics {\sem{drl}} over ${\cal D}=\bigcup_{j\in[0,5]}{{\cal D}_{n_j}}$ (red) and ${\cal D}^r$ (blue) by varying $\gamma$. Bullet styles match the legend shown on the left.}
    \label{fig:exp}
\end{figure}

Figure~\ref{fig:exp} (left), reporting the distance between the initial and final strength of the goal argument (i.e., $|\rho(\goal)-\tau_{\goal}|$), shows that three of the previous semantics considered in this paper (i.e., $\sem{mlp}, \sem{reb}, \sem{qen}$), exhibit a flat behavior, that is all of them fail to capture the expected variation in the final strength of $\goal$ with respect to $n$.  
In contrast, all the proposed semantics (dashed lines) as well as \sem{dfq} semantics exhibit a sloping descending trend of the distance $|\rho(\goal)-\tau_{\goal}|$.
Although \sem{dfq} somewhat complies with the behavior of the proposed semantics, it is important to note that it fails in satisfying several postulates of Table~\ref{tab:semantics-principles} and, in some cases, the provided results are counterintuitive (see Example \ref{ex:prel-DFQUAD}).

Among $\sem{drl}$ and $\sem{mqe}$ semantics, the latter allows the distance $|\rho(\goal)-\tau_{\goal}|$ to diminish more rapidly; 
for instance, for $n\text{=}100$, this distance is close to zero. 
Moreover, $\delta_{\mathrm{sum}}$ consistently yields lower values for $|\rho(\goal)-\tau_{\goal}|$ compared to  $\delta_{\mathrm{max}}$, due to the larger denominator, regardless of whether the semantics is $\sem{mqe}$ or $\sem{drl}$. 
In fact, for $n\in\{1,2,5,10,100,1000\}$, 
{$\delta_{\mathrm{sum}}(\goal)$ (resp., $\delta_{\mathrm{max}}(\goal)$)} reaches the value of 
{$\{\-1, \-0.66, \-0.33, \-0.18, \-0.02, \-0.002\}$} (resp., {$\{\-1.33, \-1, \-0.57, -0.33, \-0.04, \-0.004\}$})---that is, the higher $n$ the lower {$|\delta_{\agg}|$}, with  $q \in \{ \mathrm{sum},\mathrm{max}\}$.

\noindent
\textbf{Sensitivity Analysis.}\ To assess the sensitivity of \sem{drl} semantics with respect to its parameter ${\gamma}$, we conducted a parameter sweep by varying it between 0 and 3 with step-size 0.25. 
Recall that higher values of $\gamma$ emphasize the impact of attackers and supporters on the final strengths; conversely, lower values of $\gamma$ cause the strengths to be more heavily influenced by their initial values.
For $\gamma = 1$, both the initial strength and those of attackers and supporters are equally important.
These results may serve as a practical guide for selecting $\gamma$. 
Thus, we run experiments 
on the  (above presented) dataset $\cal D$ and a new dataset ${\cal D}^r = \{  \QBAF_1^{r},\dots,\QBAF_{100}^{r} \}$, that is a dataset containing $100$ distinct acyclic {\nQBAF}s $\Delta^{r}_{1},\dots, \Delta^{r}_{100}$ built as follows.
For each \nQBAF in ${\cal D}^r$ we randomly select: $i)$ the number of arguments in $[30,100]$, $ii)$ edge density (i.e., the ratio between the edges present in the graph and the maximum number of edges that the graph can contain) in $[0.1, 0.3]$, $iii)$ the ratio between the number of attackers over that of supporters in $[0.4, 0.8]$, and $iv)$ the initial arguments' score in $[0,1]$.

The results, shown in Figure~\ref{fig:exp} (right), confirm that the lower the $\gamma$, the lower the $|\rho(\goal)-\tau_{\goal}|$.

\begin{figure}[t]
    \centering
    \includegraphics[width=0.99\columnwidth]{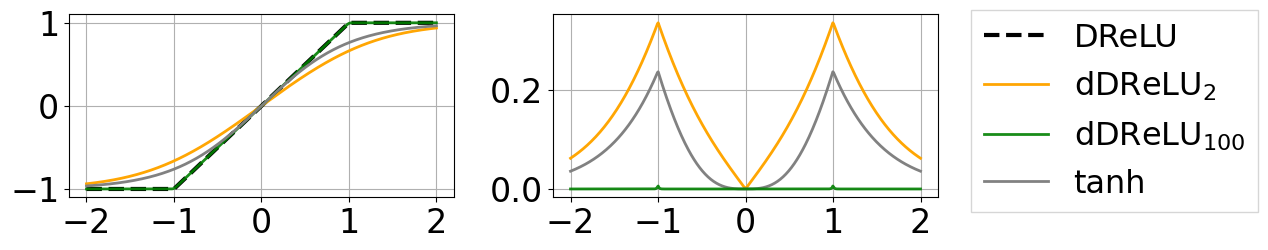}
    \vspace*{-3mm}
    \caption{(Left) Functions $\drelu$ (black dashed line) and  $f\in \{\ddrelu_{k=2},$ $\ddrelu_{k=100},$ $\funz{tanh}\}$ (orange, green, and gray colored lines). (Right) Approximation error $|\drelu - f|$.}\label{fig:grafico}
\end{figure}

\section{Convergence Analysis}\label{sec:convergence}
Given a \nQBAF and an argument $a\in\arg$, we are interested in the existence of the limit: 
$\lim\limits_{i\rightarrow \infty} \rho^i(a)$, 
where $\rho^0(a)=\tau_a$ and $\rho^j(a) = \rho(\rho^{j-1}(a))$. 
If the limit exists, we say that the computation of $\rho$ converges.
The following proposition states that, in the absence of cycles, the computation is guaranteed to converge and does so within linear time, reflecting the stability and tractability of acyclic dependency structures.

\begin{proposition}\label{prop:conv-acyclic}
    The computation of $\rho_\agg$, with $\agg\in\{\mathrm{sum},$ $\mathrm{max}\}$, under both semantics \sem{drl} and \sem{mqe}, converges in linear time for acyclic \nQBAF{s}.
\end{proposition}

\begin{proposition}\label{prop:divergence-general}
The computation of $\rho_\agg$, with $\agg\in\{\mathrm{sum},$ $\mathrm{max}\}$, under both semantics \sem{drl} and \sem{mqe}, may diverge for general \nQBAF{s}.
\end{proposition}

\begin{figure}[t]
\centering
\setlength{\tabcolsep}{1pt}
\begin{tabular}{@{}c@{}c@{}c@{}}
\begin{minipage}{0.33\columnwidth}
    \centering
    \includegraphics[width=\columnwidth]{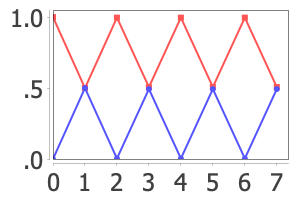}
\end{minipage}
&
\begin{minipage}{0.33\columnwidth}
    \centering
    \includegraphics[width=\columnwidth]{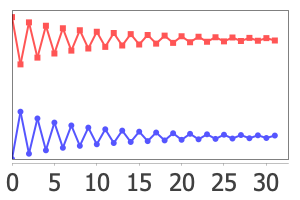}
\end{minipage}
&
\begin{minipage}{0.33\columnwidth}
    \centering
    \includegraphics[width=\columnwidth]{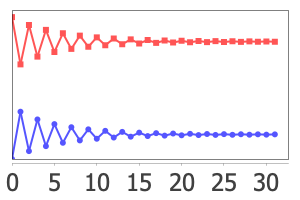}
\end{minipage}
\end{tabular}
\vspace*{-3mm}
\caption{Strength of arguments $\tt b_1$ (red) and $\tt c_1$ (blue) from the \nQBAF of  Figure~\ref{fig:unintuitive} (right) over number of iterations under  semantics {{$\sem{drl}$} and} {$\sem{ddrl}$} with {$\gamma\mbox{=}1$ and $\agg\!\in\!  \{\mathrm{sum}, \mathrm{max}\}$} (left); and with {$\gamma\mbox{=}2/3$} under semantics {\sem{ddrl} with $\agg\mbox{=} \mathrm{sum}$} (center) and $\agg\mbox{=} \mathrm{max}$  (right).
}
\label{fig:grafici}
\end{figure}

As an example, for the \nQBAF shown in Figure~\ref{fig:unintuitive} (right), the strength of argument $\tt b_1$ (resp.,  $\tt c_1$) under semantics \sem{drl} with {$\gamma=1$}, systematically alternates between $\tt 1$ (resp., $\tt 0$) and $\tt 0.5$, as shown in Figure~\ref{fig:grafici} (left).

We now move our attention to ensure more general convergence guarantees, starting from semantics $\sem{drl}$. 
To achieve this, as it will be clarified in what follows, we need to make \drelu differentiable. 
For that, as \drelu is not differentiable in $\pm 1$, we use ${\ddrelu}_k(z)$ to denote the differentiable version of \text{\drelu}\!, where $k$ is a sufficiently large positive constant, defined as follows:
\vspace*{-2mm}
\begin{equation}\label{eq:ddrelu}
\ddrelu_k(z)=\frac{1}{k}\ln\left(\frac{1+e^{k(z+1)}}{1+e^{k(z-1)}}\right)-1.
\end{equation}

Next, we characterize the properties of the function $\ddrelu_k$ as monotonicity and smoothness.

\begin{theorem}\label{prop:char}
For any $k\!\geq\! 1$, $\ddrelu_k$ is an odd function, {strictly} monotone, ascending, and infinitely differentiable.
\end{theorem}

Intuitively, $\ddrelu_k$ approximates $\drelu$ with an error that is inversely correlated to parameter $k$. 
That is, the higher parameter $k$, the better the approximation, as shown in Figure~\ref{fig:grafico} and formalized next.

\begin{proposition}\label{prop:error} 
For any $z\in \mathbb{R}$, it holds that \\
\hspace*{15mm} $\lim\limits_{k\rightarrow \infty} \ddrelu_k(z) = \drelu(z)$.
\end{proposition}

It can be observed that an approximation of the \drelu using the hyperbolic tangent would have resulted in a larger approximation error (see the gray line in Figure~\ref{fig:grafico}, right).
Moreover, the approximation error depends on $1/k$. 
Henceforth, we assume $k$ to be sufficiently large, and we denote with $\ddrelu$ the function $\ddrelu_k$ with $k=100$, which guarantees an approximation error below $10^{-3}$. 

We are ready to adapt the \sem{drl} semantics to the $\ddrelu$.
The novel semantics, denoted as $\sem{ddrl}$, is defined by replacing in Equation~\ref{eq:oursacyclic} the function \funz{DReLU} with \funz{dDReLU}.

\begin{example}\label{ex:ddrelu-discrete-1}\styleex 
    Continuing with Example \ref{ex:drelu-discrete-1}, the final strength under $\sem{ddrl}$ semantics are $\rho_{\mathrm{max}}({\goal}) =\tt 0.833$ and $\rho_{\mathrm{sum}}({\goal}) =\tt 0.909$, coinciding with those of semantics $\sem{drl}$ in Example~\ref{ex:drelu-discrete-1}.
    Consider now the \nQBAF in Figure~\ref{fig:unintuitive} (right), the strengths of arguments $\tt {b_1}$ and $\tt c_1$ under {\sem{ddrl}} semantics (with $\gamma=1$) still alternates their values (between $1$ and $0.5$ for $\mathtt{b_1}$, and between $0$ and $0.5$ for $\mathtt{c_1}$), as shown in Figure~\ref{fig:grafici} (left).\hfill$\Box$
\end{example}

Note that {semantics} \sem{ddrl} satisfies the same principles mentioned in Proposition \ref{prop:principles} for the \sem{drl} semantics. 

We next state our main convergence result for ${\sem{ddrl}}$, depending on the parameter $\gamma$ and the \nQBAF topology.

\begin{theorem}\label{thm:convergence}
Let $\QBAF=\tQBAF$ be a \nQBAF. Then, 
the computation of $\rho_\agg$, with $\agg=\mathrm{sum}$ (resp., $\agg=\mathrm{max}$), under semantics \sem{ddrl} converges if $\gamma<{\frac{2}{3d}}$ (resp., $\gamma<{\frac{1}{d}}$), where $d=\max\limits_{a\in \arg} |(\att\cup\supp)\cap(\arg\times\{a\})|$.
\end{theorem}

\begin{example}\label{ex:gamma}\styleex
    Consider again the \nQBAF in Figure~\ref{fig:unintuitive} (right) where $d=2$, and assume $\gamma={1}$. 
    According to Theorem~\ref{thm:convergence}, $\rho_{\mathrm{sum}}$ (resp., $\rho_{\mathrm{max}}$) does not converge as {${\gamma}\not< 1/3$} (resp., {${\gamma}\not< 1/2$}), as also shown in Figure~\ref{fig:grafici} (left).    
    In contrast, setting {$\gamma<1/3$} (resp., {$\gamma<1/2$}) ensures that the computation of $\rho_{\mathrm{sum}}$ (resp., $\rho_{\mathrm{max}}$) converges under \sem{ddrl} semantics.\hfill~$\Box$
\end{example}

It is worth noting that the result of Theorem~\ref{thm:convergence} provides a sufficient condition for convergence, rather than a necessary one. 
Consequently, convergence may still occur in cases where this condition is not strictly satisfied. 
As an example, when choosing $\gamma= 2/3$ in the scenario of Example~\ref{ex:gamma}, we have that \sem{ddrl} semantics converges, as shown in Figure~\ref{fig:grafici} (center and right, respectively). 
Another aspect to investigate is identifying special topologies of the underlying graph that guarantee convergence.
As stated next, convergence is guaranteed when every argument occurs in at most one cycle.

\begin{theorem}\label{thm:convergence-scc} 
     Let $\QBAF=\tQBAF$ be a \nQBAF. 
     If every argument $a\in \arg$ belongs to at most one cycle in the underlying graph $\<\arg,\att,\supp\>$, then the computation of $\rho_{\agg}$ under
     semantics $\sem{ddrl}$ converges for any     
     $\agg\in \{\mathrm{sum},\mathrm{max}\}$ and ${\gamma \in \mathbb{R}_{\geq 0}}$.
\end{theorem}

The investigation of further topological QBAF classes that guarantee convergence is deferred to future work.

Regarding theoretical convergence guarantee for $\sem{mqe}$, similarly to that of Theorem~\ref{thm:convergence}, we observe that those proposed in~\cite{Potyka19} for $\sem{qen}$ (cf. Corollary 3.5) still applies for \sem{mqe} semantics by replacing the Lipschitz constant of $\alpha$ (i.e., $L_{\alpha}=d$) with that of $\delta_{\agg}$ (i.e., $L_{\delta_{\mathrm{sum}}}=3d$ and $L_{\delta_{\mathrm{sum}}}=2d$).

\begin{figure}[t]
    \centering
\includegraphics[width=0.95\columnwidth]{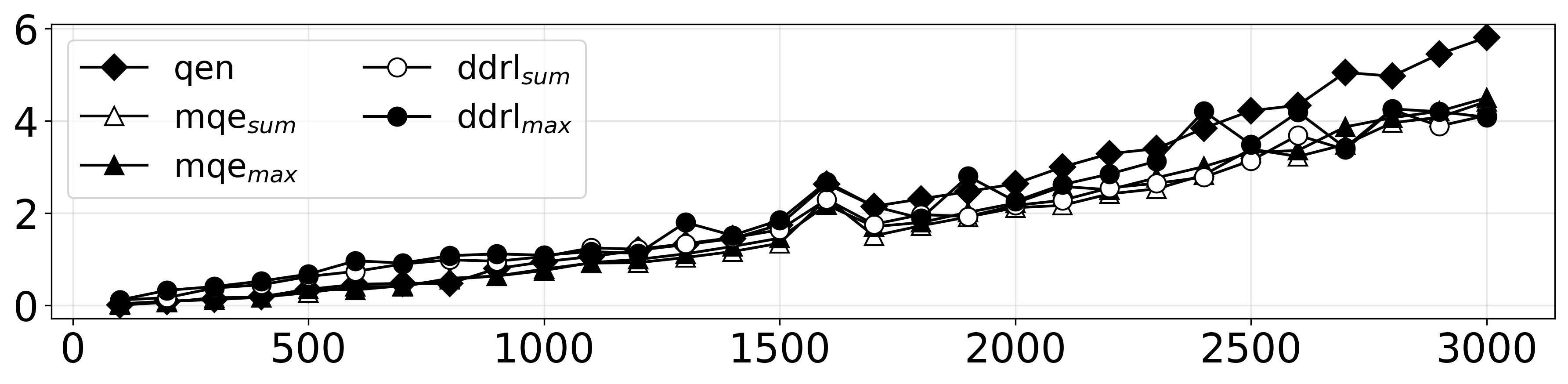}
\vspace*{-3mm}\caption{Average runtime (ms) needed for convergence over  {\nQBAF}s with $n$ arguments, where $n \in \{100, 200, ...3000\}$.\vspace*{-1mm}}
\label{fig:runtime}
\end{figure}

To investigate the practical convergence rate of the proposed semantics, particularly when assigning equal weights to the initial scores and the influence of attackers and supporters (i.e., $\gamma = {1}$), we implemented the  
proposed semantics within the \textit{Attractor Library}, and evaluated their performance using the benchmark dataset established for quadratic energy $\sem{qen}$~\cite{dataset}.
The dataset consists of $30$ groups, each containing $100$ {\nQBAF}s with $n \in \{100, 200,..,3000\}$ arguments.
Initial strengths are randomly assigned.
Our experiments show that all the proposed semantics converge for at least $95\%$ of the tested {\nQBAF}s, with runtimes scaling sublinearly \wrt the \nQBAF size (denoted by $n$). 
Figure~\ref{fig:runtime} reports the average runtime required for convergence in each group.

\vspace*{-2mm}

\section{Discussion and Conclusion}
To guarantee convergence, several modular semantics can scale
the influence of the aggregated value down by a constant value. 
However, this limits the semantics' ability to adapt the initial weight~\cite{Potyka19}, or equivalently, to reduce the `open-mindedness' principle. Intuitively, the higher the aggregated value is scaled down, the lower the possibility of $\tau_a$ being arbitrarily close to bounds $[0,1]$.
As an example, \sem{reb} is not open-minded since it does not admit final strength values smaller than $\tau_a^2$.
This is also called \textit{conservativeness} of gradual semantics~\cite{Potyka19}.

Another approach to improve convergence guarantees, alternative to the previous one of making the semantics more conservative, is to let the continuous change of arguments' strength $\rho$ be described by means of differential equations, whose unique solution is a `continuous' model $\tilde{\rho}$.
Then, $\tilde{\rho}$ is approximated (e.g., by a first-order Taylor approximation) to compute the value at each time step~\cite{Potyka19}. As $\sem{ddrl}_\agg$ and $\sem{mqe}_\agg$ are differentiable, we have implemented such an approach in the Attractor Java Library~\cite{potyka2018tutorial}. 

Other research has explored variations where argument strengths lie in $[\pm 1]$ \cite{AmgoudCLL08}, where the edges in the \nQBAF are weighted \cite{Potyka21MLP}, or where the target of a relationship is generalized to include relations, not only arguments~\cite{AmgoudDL24}. 
Although we have not addressed this in the present work, our semantics can be readily extended to accommodate weights in the interval $[\pm 1]$ (by deleting the stretching and compression steps).

Gradual semantics for structured argumentation frameworks started to gain attention~\cite{KR2025-GradualABA,skiba2023ranking,heyninck2023ranking,amgoud2015argumentation}.
Our semantics can also be adapted to cope with structured argumentation, as recently done for Assumption-based Argumentation in~\cite{KR2025-GradualABA}.

Future work will also explore neuro-symbolic approaches that use Large Language Models for the construction of (QB)AFs, as recently done in~\cite{FreedmanAAAI25,gorur-etal-2025-large,cabessa-etal-2025-argument,mayer2020ecai}.

\bibliographystyle{named} 
\bibliography{main}
\end{document}